%% file: elsarticle-template.tex
\documentclass[review]{elsarticle}

\usepackage{lineno,hyperref}
\usepackage{comment}
\usepackage[ruled,vlined]{algorithm2e}
\usepackage{amsmath,amssymb,amsfonts}
\usepackage{graphicx}
\usepackage{lscape}
\usepackage{float}

\newcommand{\etal}{\textit{et al.}}
\usepackage{subcaption}
\usepackage{multirow}
\usepackage{multicol}
\usepackage[noend]{algpseudocode}
\usepackage{booktabs}
\usepackage{verbatim}
\usepackage{color}

\modulolinenumbers[5]

\journal{Decision Support Systems}

\biboptions{authoryear}

\begin{document}

\begin{frontmatter}

\title{Evaluating Post-hoc Interpretability with Intrinsic Interpretability}

\fntext[myfootnote]{This work was supported in part by the FCT Research Grant SFRH/BD/136786/2018 and partial developed in the Hermes AI Excellence Center by the Leitat Technological Center.}

\author[cisuc,ipo]{Jos\'e P. Amorim\corref{mycorrespondingauthor}}
\ead{jpamorim@dei.uc.pt}

\author[cisuc]{Pedro H. Abreu}
\ead{pha@dei.uc.pt}

\author[ipo]{Jo\~ao Santos}
\ead{joao.santos@ipoporto.min-saude.pt}

\author[HES-SO]{Henning M\"uller}
\ead{henning.mueller@hevs.ch}

\address[cisuc]{CISUC, Department of Informatics Engineering, University of Coimbra, Portugal}
\address[ipo]{IPO-Porto Research Centre, Porto, Portugal}
\address[HES-SO]{University of Applied Sciences Western Switzerland (HES-SO Valais), Sierre, Switzerland}

\cortext[mycorrespondingauthor]{Corresponding author}

\begin{abstract}
Despite Convolutional Neural Networks having reached human-level performance in some medical tasks, their clinical use has been hindered by their lack of interpretability. Two major interpretability strategies have been proposed to tackle this problem:  post-hoc methods and intrinsic methods.
Although there are several post-hoc methods to interpret DL models, there is significant variation between the explanations provided by each method, and it a difficult to validate them due to the lack of ground-truth.
To address this challenge, we adapted the intrinsical interpretable ProtoPNet for the context of histopathology imaging and compared the attribution maps produced by it and the saliency maps made by post-hoc methods. To evaluate the similarity between saliency map methods and attribution maps we adapted 10 saliency metrics from the saliency model literature, and used the breast cancer metastases detection dataset PatchCamelyon with 327,680 patches of histopathological images of sentinel lymph node sections to validate the proposed approach.
Overall, SmoothGrad and Occlusion were found to have a statistically bigger overlap with ProtoPNet  while Deconvolution and Lime have been found to have the least.
\end{abstract}


\end{frontmatter}

%
\section{Introduction}\label{sec:Intro}
Despite Convolutional Neural Networks (CNN) having reached human-level performance in lymph node metasteses detection from retrospective histopathological images~\cite{Bejnordi2017}, their inherent lack of interpretability remains a crucial disadvantage when compared with other traditional classification methods for their use in a clinical setting.
Interpretability is the capability of assigning meaning to an instance by a translation that does not change its original validity~\cite{Graziani2022}. A possible translation may be to move to a simplified space that is easier for humans to understand.

Over the last few years, a vast number of interpretability methods~\cite{Molnar2019book} have been proposed for explaining CNN predictions. These methods can be divided in: intrinsic interpretability, referring to models that can be explained without further methods, usually by restricting the model's complexity; and post-hoc interpretability, referring to methods that explain the model decisions after training and without influencing the model itself.

Saliency maps are one of the most popular post-hoc interpretability methods to explain CNNs in the context of medical imaging~\cite{Amorim2022}. This strategy illustrates the importance of individual pixels of the input image on the overall prediction of a CNN. The color or intensity of each pixel corresponds to the weight that the same pixel in the input image had on the classification process. Even though most methods have been shown continuously to be able to highlight regions with relevant medical evidence, the saliency maps generated by different methods exhibit a significant degree of variation, which is evidence of bias specific to each method that can not be overlooked. The qualitative analysis and evaluation of saliency map methods remain an open challenge.

Several intrinsic interpretability strategies have also been adopted in medical imaging~\cite{Amorim2022} with the objective of constraining the behavior of the classification model making it understandable to humans. Case-based reasoning closely approximates the reasoning process of physicians, who take decisions based on similar cases. Prototypes are special cases of case-based reasoning, where a small number of data points can represent the entire dataset~\cite{Molnar2019book}. 
To this end, ProtoPNet~\cite{Cynthia2019}, a deep learning architecture based on a convolutional neural network, learns the optimal prototypes automatically and makes predictions based on the similarity of the instance to each prototype.

This network also generates an attribution map (i.e. heatmap) that highlights the location in the image which closely resembles the prototypes. 

While both saliency map methods and ProtoPNet are capable of highlighting the importance of a region in the image for the prediction of the network, they do this through different mechanisms. 

As intrinsic interpretability methods are considered to be more faithful to the underlying model's behavior~\cite{Cynthia2019} as they do not require an external method after training, it can become a good ground-truth of what regions of the image the saliency maps should focus.
Therefore, we propose in this research work to use the intrinsic interpretable model's explanation as ground-truth and measure the overlap between it and the post-hoc explanations to validate them.
Therefore, we propose an approach that automatically validates the results generated by post-hoc methods by comparing their results overlapped with the results achieved by an intrinsic interpretable method that is used as ground-truth.
To this end, we adapted ProtoPNet for digital pathology and evaluated the overlap between it and 8 different saliency map methods using 10 saliency metrics. 

In our experimental setup, we have trained three CNNs and three ProtoPNets based on three architectures (ResNet18, ResNet152 and DenseNet101) on the PatchCamelyon dataset (histopathologic scans of lymph node sections). Following the training of the networks, 8 different saliency map methods were used to extract saliency maps of the test set to be compared with attribution maps generated intrinsically by the ProtoPNets. To evaluate this approach, 10 metrics adapted for evaluating saliency maps~\cite{Bylinskii2019} were used.

Overall, despite some variation in results depending on the CNN architecture used, SmoothGrad and Occlusion have been found to be statistically more similar to ProtoPNet, while Deconvolution and Lime have been found to be more dissimilar.

This article follows the following structure: Section~\ref{sec:Background} will briefly review related works in the literature. In Section~\ref{sec:Method} we present the different components of the study for evaluating the connection between saliency map methods and prototypes activation maps: 1) data selection, 2) model training, 3) prototypical parts extraction 4) saliency map extraction, and 5) saliency map evaluation. We present and discuss
the results of the experiments in Section~\ref{sec:Results}. Finally, in Section~\ref{sec:Conclusions} we conclude with our final remarks and steps for future work.

\section{Background}\label{sec:Background}
Interpretability is an important prerequisite for the adoption of computer-aided diagnosis systems.
To this end, different computation competitions have emerged, and one of those is the Camelyon16 challenge~\cite{Bejnordi2017} has the goal of evaluating algorithms on the task of automatic detection of breast cancer metastases in whole-slide images of hematoxylin and eosin (H\&E) lymph node sections. Convolutional neural networks (CNNs) have been successful in this task, with approaches achieving area under the receiver operating curve (AUC) of 0.925, increasing to 0.995 when combined with pathologists' diagnosis (approximately 85 percent reduction in human error rate). 
But, although the potential is shown, without a clear understanding of their reasoning process, their application in clinical practice remains elusive. 

There are two distinct approaches for this problem: intrinsic interpretability and post-hoc interpretability.
Among post-hoc interpretability methods, the most adopted in medical imaging are saliency map methods. This can be especially seen in the oncological field~\cite{Amorim2022} where half of the interpretability strategies employed to understand deep learning models are saliency map methods. These methods can be divided into two major groups: back-propagation methods and occlusion or sensitivity methods~\cite{Molnar2019book}. 

While not as popular, intrinsic interpretability strategies have also been adopted in medical imaging~\cite{Amorim2022}. The first is the approximation of the network with an intrinsically interpretable model (i.e. decision rules) of similar performance but easier to understand~\cite{Amorim2018}.
Rather than generating saliency maps after training, it is also possible to incorporate in the network the heatmap generation through an attention mechanism or probability estimation (i.e. pixel-wise or patch-wise). The patch-wise heatmap has shown remarkable results in the diagnosis malignancy-based dermoscopic images~\cite{Radhakrishnan2017}.
Text explanations have become a reality with the advent of language models. An example is the training of a language model alongside the visual model for extracting text explanations while classifying the malignancy of mammograms Lee~\etal~\cite{Lee2019}.

Case-based reasoning is another intrinsic interpretability strategy that closely approximates the reasoning process of physicians, as they have to extract from their knowledge acquired from looking at similar cases. In this strategy, the classification of an instance is based on the classes of similar instances~\cite{Cynthia2021}.
Prototypes are special cases of case-based reasoning, where a small number of data points are selected to represent all the data. 
Prediction of a data point can then be made by their similarity and dissimilarity with prototypes of either class.
ProtoPNet~\cite{Cynthia2019} learns automatically the optimal prototypes in the data and makes predictions based on the similarity of the features extracted by a CNN and the features of the prototypes. With this added prototype layer, the network is capable of explaining the prediction based on a similarity score to each prototype and a heatmap denoting the location of each prototypical part. While ProtoPNet was not applied to medical imaging, an extension called IAIA-BL\cite{Cynthia2021} which adds a component of fine annotation has shown great results in the classification of mammograms. 

Although both saliency maps extracted via post-hoc interpretability methods (i.e. back-propagation) and attribution maps extracted via intrinsic interpretability methods (i.e. ProtoPNet) both highlight the important regions for the classification, they do this fundamentally through different mechanisms. 
Also, intrinsic interpretability methods are perceived as more trustworthy and more faithful to the behavior of the underlying model than post-hoc interpretability methods~\cite{Cynthia2019}.

The evaluation of saliency maps is difficult because there is a lack of ground-truth on how the ideal saliency map should look. One strategy to evaluate a saliency map is by looking at the drop in confidence in the prediction when obscuring a region highlighted by the saliency map~\cite{Samek2017, Alvarez-Melis2017}. This strategy demonstrated, in the digital pathology context, that reducing the complexity of the network had a positive impact on how much the saliency maps produced by the network reflected the model's reasoning~\cite{Amorim2020}.

Due to the fact of the higher trustworthiness of intrinsic interpretability methods, the attribution maps provided by ProtoPNet can be used as a ground-truth, allowing saliency map methods that produce saliency maps with a bigger overlap to the ground-truth to be considered more faithful to the model. 
Therefore, we developed an approach to evaluate saliency maps by overlapping the saliency map produced by a post-hoc method and the attribution map produced by an intrinsic methods 

But, as we want to compare the heatmap generated by a saliency map method and prototypical part network, we can select the latter as our ground-truth and measure the overlap between them.
Thus, we can use different metrics used for evaluating saliency models which produce heatmaps representing the probability of an individual looking at the pixel.

\section{Method}\label{sec:Method}

\begin{figure}[!ht]  
    \centering
    \includegraphics[width=0.7\textwidth]{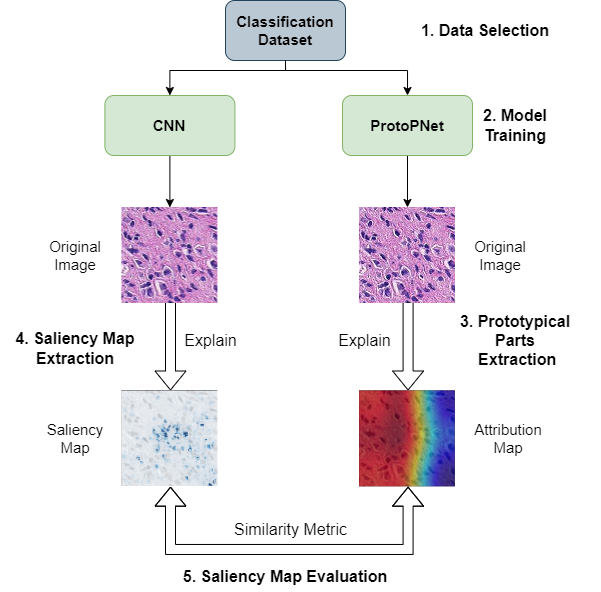}
    \caption{Automatic evaluation of saliency maps perturbation pipeline.}
    \label{fig:PREPRINT_method}
\end{figure}

Having in mind the main goal of this work which consists in comparing the saliency map methods and prototypical parts activation maps, a five-stage pipeline was defined in the experimental setup and is illustrated in Figure~\ref{fig:PREPRINT_method}: 1) data selection, 2) model training, 3) prototypical parts extraction 4) saliency map extraction, and 5) saliency map evaluation.

\subsection{Data Selection}
To evaluate the proposed approach, the PatchCamelyon dataset~\cite{Veeling2018} derived from the Camelyon16 dataset\cite{Bejnordi2017} was used. The Camelyon16 dataset contains 400 H\&E stained whole-slide images of sentinel lymph node sections split into 270 slides with pixel-level annotations for training and 130 unlabeled slides for testing.
PCam dataset contains 327,680 patches with size of 96 x 96 pixels extracted from Camelyon16 with a 10x magnification. The PatchCamelyon task is to classify the images into benign or malignant cases based on expert segmentations of malignant tissue.

PatchCamelyon dataset was chosen because of the quality of the images which were curated and segmented by pathologists and the large number of images.

The patches were first normalized into fixed range between 0 to 1 to improve the optimization process. Also to avoid overfitting, data augmentation was applied to increase diversity of the images of the training set. Images were randomly flipped both vertically and horizontally, and random brightness augmentation was used.

\subsection{Model Training}

Three CNN architectures were explored: Resnet18~\cite{He2015}, Resnet152~\cite{He2015} and DenseNet101~\cite{Huang2017}. These CNN architectures were chosen as they represent state-of-the-art approaches for many medical imaging tasks and they achieved good results in metastasis detection in the Camelyon16 challenge~\cite{Bejnordi2017}.

For the purpose of achieving the best performance on the medical imaging task of tumor detection of the PatchCamelyon dataset, the three CNNs were pre-trained on the ImageNet dataset~\cite{ImageNet} and fine-tuned on the medical dataset with a low learning rate.

All the models were trained with a batch size of 64 images, for 100 maximum epochs which were cut short by stoping the training early when the validation accuracy stops improving. The learning rate also was reduced by a factor of 0.2 when the validation accuracy also plateaus.

Hyperparameter optimization through grid search was used to select thee optimal optimization algorithm and initial learning rate.

\subsection{Prototypical Part Network}

\begin{figure}[!ht]  
    \centering
    \includegraphics[width=0.9\textwidth]{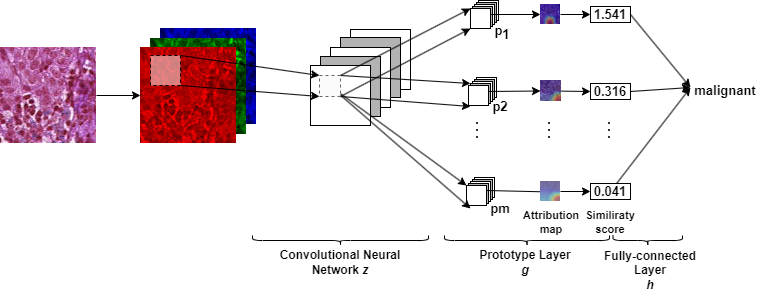}
    \caption{Architecture of ProtoPNet.}
    \label{fig:PREPRINT_protopnet}
\end{figure}

The ProtoPNet network~\cite{Cynthia2019} (Figure~\ref{fig:PREPRINT_protopnet}) is composed of a convolutional neural network that extracts features for the classification $z = f(x)$, followed by a prototype layer $g$, and a fully connected layer $h$. The CNN component is based on the three CNN architectures mentioned previously. 

The prototype layer learns $m$ prototypes which can represent the entire training set. Each prototype represents a different prototypical part or concept either of the malignant or benign class. After learning the prototypes, the layer  computes the similarity scores between all patches of the image of the same size as the prototype using the $L^2$ distance function.
The result is an attribution map for each prototype that indicates the regions of the image where it's most represented.

On the original ProtoPNet paper~\cite{Cynthia2019}, the attribution map is created by aggregating the similarity scores using global max pooling. Rather than using max pooling, top-k average pooling was used~\cite{Cynthia2021} to use top 5\% of the most activated convolutional patches that are closest to each prototype, instead of only the top most activated patch.

Finally, the similarity scores produced by the prototype layer are multiplied by the weight matrix of the fully connected layer to produce the output logits, which when passed through the softmax function produce the predicted probabilities for each class.

\subsubsection{ProtoPNet training algorithm}
Training of ProtoPNet is divided into three phases: (1) stochastic gradient descent (SGD) of layers before the last layer; (2) projection of prototypes; (3) convex optimization of the last layer.

In the first training stage, the convolutional layers' parameters and the prototype layer's parameters are optimized while keeping the last layer fixed.

During this phase, the loss function minimized is composed of the weighted sum of three losses: cross-entropy loss (CrsEnt), the cluster cost (Clst), and separation cost (Sep) (Equation~\ref{eq:PREPRINT_loss}):

\begin{equation}\label{eq:PREPRINT_loss}
\min \lambda_1 \text{CrsEnt} + \lambda_2 \text{Clst}+\lambda_3 \text{Sep}
\end{equation}

The cross-entropy loss (CrsEnt) encourages the predicted classes to be the same as the target $y_i$ in the training set composed of $n$ instances (Equation~\ref{eq:PREPRINT_CrsEnt}):

\begin{equation}\label{eq:PREPRINT_CrsEnt}
\text{CrsEnt}=  \frac{1}{n} \sum_{i=1}^n \text{CrsEnt}\left(h \circ g_{p} \circ f\left(x_i\right), y_{i}\right)
\end{equation}

The cluster cost (Clst) encourages each training image to have some latent patch that is close to at least one prototype ($p_i \in P$) of its own class (Equation~\ref{eq:PREPRINT_clst}):

\begin{equation}\label{eq:PREPRINT_clst}
\text{Clst}=\frac{1}{n} \sum_{i=1}^n \min _{j: p_j \in P_{y_i}} \min _{z \in \operatorname{patches}\left(f\left(x_i\right)\right)}\left\|z-p_j\right\|_2^2
\end{equation}

The separation cost (Sep) encourages every latent patch of a training image to stay away from the prototypes not of its own class (Equation~\ref{eq:PREPRINT_sep}):

\begin{equation}\label{eq:PREPRINT_sep}
\text{Sep}=-\frac{1}{n} \sum_{i=1}^n \min _{j: p_j \notin P_{y_i}} \min _{z \in \operatorname{patches}\left(f\left(x_i\right)\right)}\left\|z-p_j\right\|_2^2
\end{equation}

During the second phase of the projection of prototypes, each prototype is projected onto the nearest training image patch from the same class as the prototype. This is done so that when interpreting the predictions made by the network, the prototypes represent actual patches of images in the training set.

Finally, in the last phase, the last layer is optimized using a convex optimization focusing on a sparsity property making the model rely more on positive evidence (i.e. predicting a class by using the prototypes of that class) and rely less on negative evidence (i.e. prototypes from the negative classes). 

To find the optimal hyperparameters for each ProtoPNet architecture, grid search  algorithm was used. The hyperparameters that were optimized were the number of prototypes, dimensions of prototypes, learning rates in each training phase, and weights for the three components of the loss functions - cross-entropy loss, cluster cost, and separation cost.

\subsection{Saliency Map Methods}

We have selected 8 popular saliency map methods of the two major groups: back-propagation methods and occlusion or sensitivity methods~\footnote{ The implementation of the saliency map methods was done using the pytorch captum toolbox~\cite{captum}.}.

Back-propagation methods compute the relevance of a pixel by propagating a signal from the output neuron backward through the layers to the input image in a single pass~\cite{Bach2015}. 
Sensitivity methods compute pixel relevance by making small changes in the pixel value of the input image and compute how the changes affect the prediction~\cite{Simonyan2014}.


\paragraph{Saliency} Saliency~\cite{Simonyan2014}, or gradient back-propagation, is a simple method where the pixel's sensitivity is given by the gradient of the loss function for the class we are interested in with respect to the input pixels. 
Each saliency map pixel's value represents how much a tiny change in the pixel would change the classification score for class c.
The gradient method generates a highly noisy saliency map.

\paragraph{Deconvolution}
Deconvolution~\cite{Zeiler2014} provides a way to map the activation of intermediate layers back to the input layer. This mapping is performed by a Deconvolutional Network which attaches to the CNN layers and performs the opposite operation. For example, the unpooling layer does the inverse of the pooling layer.

\paragraph{GuidedBackprop} GuidedBackprop~\cite{Springenberg2015} adds an additional guidance signal from the higher layers to the usual back-propagation. It combines deconvolution with back-propagation, by masking out negative values from either method.

\paragraph{SmoothGrad} SmoothGrad~\cite{Smilkov2017} is a variant of Gradient Back-propagation where the saliency map is smooth out by creating noisy copies of the input image and then average the gradient saliency maps of these noisy images. The resulting effect is a more sharp saliency map with less noisy on irrelevant regions.

\paragraph{Integraded Gradients} Integrated Gradient~\cite{Sundararajan2017} saliency map is computed by drawing a straight line in the network feature space from a baseline image and the input image and accumulating the gradients at all points along the path. The baseline image should ideally have no signal, so similarly to the original paper, we have also used a zero-based image (i.e. black image) as our baseline.

\paragraph{Occlusion Sensitivity} Occlusion Sensitivity~\cite{Zeiler2014} computes the importance of regions of the image by inspecting if there is a drop in the confidence of the model in the predicted class when the region is occluded using a mask.

\paragraph{SHAP}
Shapley Additive explanations (SHAP)~\cite{Lundberg2017} requires the training of a distinct predictive model for each distinct combination of input features. By inspecting the gap between the predictions of two predictive models when a feature is added/subtracted, we can infer the importance of the feature in the prediction. Features whose presence or absence produced a large gap in predictions have large Shapley values and are deemed important.

\paragraph{LIME}
Local Interpretable Model-agnostic Explanations (LIME)~\cite{Ribeiro2016} first produces an artificial dataset by occlusing each feature of the original datapoints. Weights are assigned to the generated datapoints based on the closeness to the original point.
Based on the generated weighted data a linear regression model is trained.
The coefficients of the linear regression correspond to the importance of the input features to the model's predictions.

\subsection{Saliency Map Evaluation}

The metrics used to evaluate saliency models were adapted for evaluating saliency maps and attribution maps~\cite{Bylinskii2019}. 
The main task of a saliency model is predicting eye movements made during image viewing. The saliency model produces a heatmap in which the pixel value represents the probability of an individual looking at the pixel. Evaluation of the saliency model consists of comparing the heatmaps to the ground-truth fixation map. We use the saliency metrics to compare the saliency map extracted from the CNNs and the attribution map of each prototype produced by the ProtoPNet.

Following Riche~\etal~\cite{Riche2013} we divided the metrics based on location-based or distribution-based and similarity or dissimilarity. This classification is summarized in Table~\ref{tab:PREPRINT_saliency_metrics}.

\begin{table}[!ht]
    \centering
    \begin{tabular}{|c|c|c|}
        \hline
        \textbf{Metrics} & \textbf{Location-based} & \textbf{Distribution-based} \\ \hline
        \textbf{Similarity} & jAUC, bAUC, sAUC, NSS, IG & SIM, CC \\ \hline
        \textbf{Dissimilarity} & MSE, MAE & KL \\ \hline
    \end{tabular}
    \caption{Saliency metrics divided by type.}
    \label{tab:PREPRINT_saliency_metrics}
\end{table}

Location-based metrics consider saliency map values as discrete locations at different threshold levels, while distribution-based metrics treat both saliency maps as continuous distributions. 

Similarity metrics measure how similar two saliency maps, while dissimilar metrics measure how dissimilar they are. Similarity should have higher values when we expect the saliency maps to not change (i.e. introduce evidence from the same class) while being lower when we expect a change (i.e. introduce evidence from a different class). The opposite should happen with the dissimilarity metrics.

\subsubsection{Location-based metrics}

Location-based metrics score saliency maps regarding how accurately they predict discrete pixel locations.

\paragraph{Area under ROC Curve (AUC)} 
The Area under the ROC curve is the most widely used metric for evaluating saliency maps. 
When computing the AUC, the saliency map is treated as a binary classifier at various threshold values and the ROC curve represents the true and false positive rates for each threshold value.

Different AUC implementations differ in
how true and false positives are calculated. 

AUC-Judd (jAUC) use a threshold level as a cut-off value to determine if pixel values in a saliency map are positives or negative.

AUC-Borji (bAUC) uses uniform random sample of image pixels as negatives and defines the saliency map values above threshold at these pixels as false positives.

Shuffled AUC (sAUC) penalizes center bias by samplying negative samples predominantly from the image center.

These saliency metrics were adapted by binarizing the ground truth saliency map by setting a threshold and selecting the most salient pixels.

\paragraph{Normalized Scanpath Saliency (NSS)} 
The Normalized Scanpath Saliency (NSS) is a similarity metric which measures the average normalized saliency map values of the locations of the ground truth saliency map.

Given a saliency map $P$ and a binarized ground truth saliency map $Q^B$, NSS can be computed so:

\begin{align} \label{eq:PREPRINT_nss}
NSS(P, Q^B) = \frac{1}{N}\sum_i\overline{P_i} \times Q_i^B \\
\text{where } N = \sum_i Q_i^B \text{ and } \overline{P} = \frac{P - \mu(P)}{\sigma(P)}
\end{align}

where $i$ indexes the i-th pixel, and N is the total number of fixated pixels.

NSS is sensitive to false positives, as the metric is normalied over all the positive pixels on the binarized ground truth saliency map.

Similar to the AUC variants, the ground truth saliency map was binarized.

\paragraph{Mean Average Error (MAE) } 
Mean Average Error (MAE) represents the average difference between the model's prediction and ground-truth (Equation~\ref{eq:PREPRINT_mae}) and can be used in regression problems. 

\begin{equation}\label{eq:PREPRINT_mae}
MAE = \frac{1}{N}\sum_{i=1}^N(\hat{y_i} - y_i)
\end{equation}

\paragraph{Mean Squared Error (MSE) } 
Mean Square Error (MSE) represents the average squared difference between the model's prediction and ground-truth (Equation~\ref{eq:PREPRINT_mse}).

\begin{equation}\label{eq:PREPRINT_mse}
MSE = \frac{1}{N}\sum_{i=1}^N(\hat{y_i} - y_i)^2
\end{equation}

\paragraph{Information Gain (IG)} 
Information Gain (IG) is a similarity information theoretic metric that measures saliency model performance compared to a baseline.

Given a binary map of pixels $Q_B$, a saliency
map $P$, and a baseline map $B$, information gain is computed as:

\begin{equation}
    IG(P, Q^B) = \frac{1}{N} \sum_i Q_i^B [log_2(e + P_i) - log_2(e + B_i)]
\end{equation}

where $i$ indexes the i-th pixel, N is the total number of fixated pixels, $e$ is for regularization, and information gain is measured in bits per fixation.

A score above zero indicates the saliency map is better than the baseline at predicting the fixated locations.

Similar to the AUC variants and Infogain, the ground truth saliency map was binarized.

\subsubsection{Distribution-based metrics}

Distributed-based metrics treats pixel values and locations of ground truth saliency maps as possible samples from an underlying distribution.

\paragraph{Similarity (SIM)} 
The similarity metric (SIM) measures the similarity between two distributions, viewed as histograms. 
SIM is computed as the sum of the minimum values at each pixel, after normalizing the input maps. Given a saliency map P and a continuous fixation map $Q^D$:

\begin{align} \label{eq:PREPRINT_sim}
SIM(P, Q^D) = \sum_i min(P_i, Q_i^D) \\
\text{where } \sum_i P_i = \sum_i Q_i^D = 1
\end{align}

iterating over discrete pixel locations $i$.

A SIM of one indicates the distributions are the same, while a SIM of zero indicates no overlap.

\paragraph{Pearson's Correlation Coefficient (CC)}
The Pearson’s Correlation Coefficient (CC) is a statistical method for measuring how correlated or dependent two variables are.
If we consider the distribution of pixels in the saliency map $Q^D$, and the saliency map $P$ as random variables, we can measure their linear relationship: 

\begin{equation}
    CC(P, Q^D) = \frac{\sigma(P,Q^D)}{\sigma(P) \times \sigma(Q^D)}
\end{equation}

where $\sigma(P, Q^D)$ is the covariance of $P$ and $Q^D$.
It is a similarity metric, which means that high positive CC values occur at locations where both the saliency map and ground truth saliency map have values of similar magnitudes.

\paragraph{Kullback-Leibler divergence (KL)}
The Kullback-Leibler divergence (KL) is a dissimilarity metric based on general information theory and it measures the difference between two probability distributions.

The KL metric takes as input a saliency map $P$ and a ground truth saliency map $Q^D$, and evaluates the loss of information when P is used to approximate $Q^D$:

\begin{equation}
    KL(P, Q^D) = \sum_i Q_i^D log(\frac{Q_i^D}{P_i})
\end{equation}

where $e$ is a regularization constant. 

One characteristic of KL is that it penalizes very sparse saliency maps.

\subsubsection{Performance Metrics}

One way to compare the predictions of the CNN and ProtoPNet is to measure the difference in confidence in the class from both models.
We have selected a number of performance metrics for evaluating a classification model, namely AUC, accuracy, precision, recall, and Mean Squared Error (MSE). These metrics are not only used to measure the performance of each model by comparing their predictions with actual labels but also to measure between predictions of CNNs and ProtoPNet with the same pre-trained base model.

True positives (TP) and true negatives (TN) represent the instances correctly classified by the model as being positive and negative, respectively. In the other-hand false positive (FP) and false negative (FN) represent the instances in which the model incorrectly classified as being positive and negative, respectively.

Accuracy represents the ratio of examples correctly classified (Equation~\ref{eq:PREPRINT_accuracy}).

\begin{equation}\label{eq:PREPRINT_accuracy}
Accuracy = \frac{TP + TN}{TP + TN + FP + FN}
\end{equation}

Precision measures which proportion of the samples predicted as positive are actually positive: 

\begin{equation}\label{eq:PREPRINT_precision}
Precision = \frac{TP}{TP + FP}
\end{equation}

Recall on the other-hand measures which proportion of the actual positive samples were collectively predicted as positive: \begin{equation}\label{eq:PREPRINT_recall}
Recall = \frac{TP}{TP + FN}
\end{equation}

In this section, we have introduced an approach for the evaluation of saliency maps using realistic perturbations which avoids the problem of creating out-of-distribution images. We have also proposed the adaptation of saliency metrics used to evaluate saliency models for the comparison of saliency maps. The proposed approach was validated by evaluating 8 saliency map methods on a digital pathology dataset called PatchCamelyon.

\section{Results}\label{sec:Results}

In the experimental setup, three CNNs pre-trained on the ImageNet dataset have been selected: RestNet18, RestNet152, and DenseNet121. We trained the three CNNs on the PatchCamelyon dataset and also used them as base models for three ProtoPNets.

Table~\ref{tab:PREPRINT_performance} shows the classification results of the models. Overall, both CNNs and ProtoPNets were able to achieve very high performance across all architectures.
The most accurate model was the ProtoPNet based on the DenseNet121 architecture with an AUC of 0.981, the corresponding CNN having achieved an AUC of 0.975. 

While the black-box CNNs perform better when using a ResNet18 and ResNet152 base model, ProtoPNet with the DenseNet121 was the best model. 

From the results, we can conclude that there was no clear trade-off of performance by adding the interpretability layer (i.e. prototypical layer) to achieve intrinsical interpretability.

\begin{table}[!ht]
\centering
\scalebox{0.9}{
\input{tables/performance}
}
\caption{Performance of models on malignancy detection.}
\label{tab:PREPRINT_performance}
\end{table}

\begin{figure}[!ht]
\makebox[\textwidth][c]{%
  \includegraphics[width=0.9\textwidth]{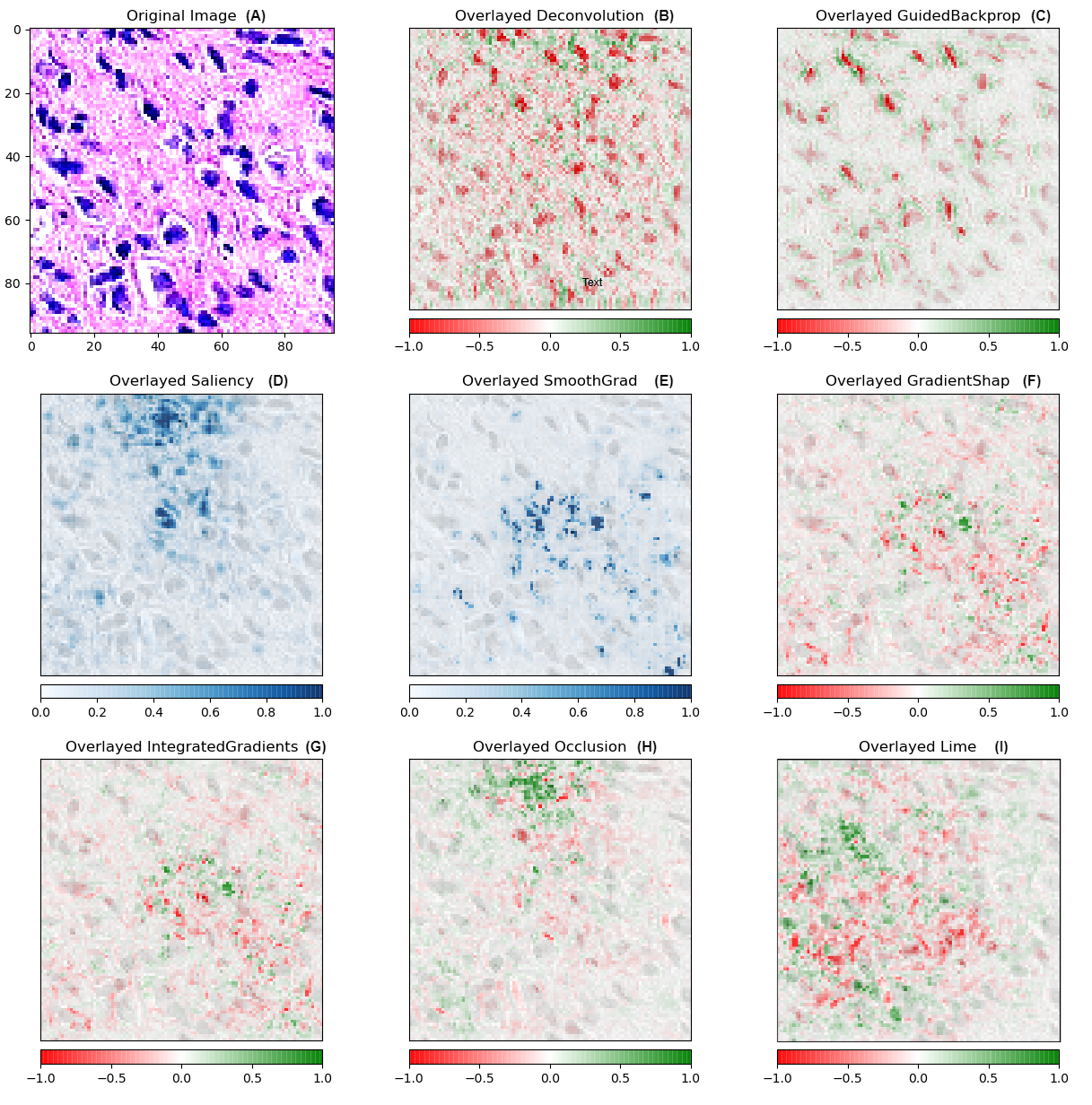}%
}
\caption[Qualitative analysis of saliency map methods for the DenseNet121 network.]{Qualitative analysis of saliency map methods for the DenseNet121 network. At the top left corner, we have the original image. The other images corresponding to the different saliency maps overlayed on top of the original image. Saliency maps with only absolute values represent positive evidence with blue tones, whereas darkness represents importance. Other saliency maps represent positive evidence of malignancy with green tones and negative evidence with red tones.}
\label{fig:PREPRINT_sm}%
\end{figure}

To qualitatively evaluate the saliency map methods we have selected one example of malignancy and computed the saliency maps for each method for the network that achieved the best results (DenseNet121). Figure~\ref{fig:PREPRINT_sm} presents side by side the result saliency maps overlapping the original image (A). Some of the methods (D and E) produce a saliency map with absolute values so the importance of each pixel is depicted by the darkness of the blue tone. Other methods (B, C, F, G, H, and I) produce both negative importance which is depicted in red, and positive importance depicted in green. Positive importance corresponds to evidence of malignancy while negative importance corresponds to evidence of benignity.

Deconvolution and Lime (B, I) highlight a vast region of the image but are able to somewhat focus on a vast number of nuclei.
In comparison, methods such as SmoothGrad, Occlusion, and GuidedBackprop (C, E, H) appear more sparse in their activations.

\begin{figure}[!ht]
\makebox[\textwidth][c]{%
  \includegraphics[width=0.7\textwidth]{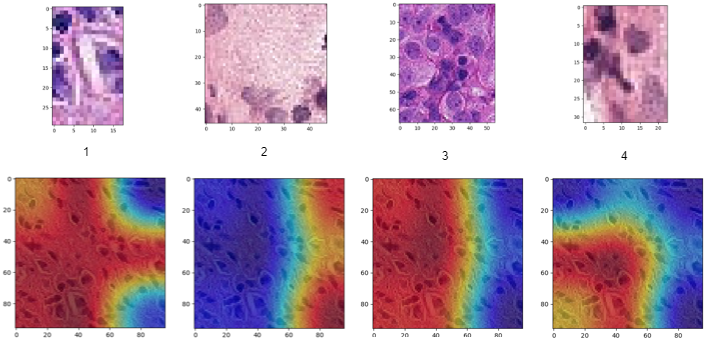}%
}
\caption[Qualitative analysis of prototypical parts of the ProtoPNet trained with the DenseNet121 architecture.]{Qualitative analysis of prototypical parts of the ProtoPNet trained with the DenseNet121 architecture. At the top are the regions of the training images that are the most similar to each of the four prototypes. At the bottom are the attribution maps depicting the similarity of the original image to each prototype. The same original image as in Figure~\ref{fig:PREPRINT_sm} was used.}
\label{fig:PREPRINT_ppnet}%
\end{figure}

To qualitatively evaluate the ProtoPNet's attribution maps we have selected the same test image as before and computed the attribution maps for 4 specific prototypes of the DenseNet121 base network. Figure~\ref{fig:PREPRINT_ppnet} presents in the top row, for each prototype its most similar region extracted from the images of the training set. Below each image is shown the attribution map calculated for the original image used before.

When compared with the previous saliency map methods, the attribution maps generated by the ProtoPNet are very soft and smooth. This can be justified by the fact that the prototype dimensions that the hyperparameter optimization chose are small which can become a trade-off between increasing performance while disregarding fine-grained explanations.

We compared the predictions of the CNNs and ProtoPNets across all three architectures on the test set to understand if both models correlate with each other. Table~\ref{tab:PREPRINT_labels} shows that AUC is high for the three architectures, suggesting that the CNN and ProtoPNet versions make similar predictions. DenseNet121, which was the most accurate architecture in the malignancy prediction task is also the architecture in which the predictions from the CNN and ProtoPNet versions are most related.

\begin{table}[!ht]
\centering
\scalebox{1}{
\input{tables/labels}
}
\caption{Comparing the predictions of CNN and ProtoPNet based on the same architecture.}
\label{tab:PREPRINT_labels}
\end{table}

To measure the overlap between CNN's saliency maps methods to ProtoPNet's attribution maps we have extracted the explanations for each image in the test set and computed the saliency metrics (10) comparing each saliency map with the corresponding attribution map.

A statistical comparison was performed using the mean of each of the 8 saliency map methods and the Friedman rank test. The results were divided into three tables for each of the three architectures.

The averaged ranks across all metrics for the DenseNet121, ResNet152 and ResNet18 architectures are shown in Table~\ref{tab:PREPRINT_average_rank}. The methods that are shown to have the smallest overlap with ProtoPNet than other methods are highlighted.

\begin{table}[!ht]
\centering
\renewcommand{\arraystretch}{0.7}
\scalebox{0.8}{
\input{tables/average_rank}
}
\caption{Average rank of similarity between saliency map methods and ProtoPNet over the 10 saliency metrics.}
\label{tab:PREPRINT_average_rank}
\end{table}

Following the work of Den\v{s}ar~\cite{Demsar2006} with $N = 10$ (number of metrics) and $k = 8$ (number of saliency map methods), the 8 methods were compared among themselves for a 5\% significance level using the two-tailed Nemenyi test ~\cite{Demsar2006} obtaining a CD (critical value for the difference of mean ranks between the 8 methods) of 2.949.

In the DenseNet121 architecture, a statistically significant difference was found between SmoothGrad and both Deconvolution and Lime on all prototypes. When taking into account only the last two prototypes, Deconvolution and Lime have been found to have a smaller overlap than the majority of other methods. By analyzing the results, methods such as SmoothGrad, Occlusion, and Saliency continuously have a bigger overlap than methods such as Deconvolution, Lime, and GradientShap.

\begin{figure}[!ht]
\makebox[\textwidth][c]{%
  \includegraphics[width=1.3\textwidth]{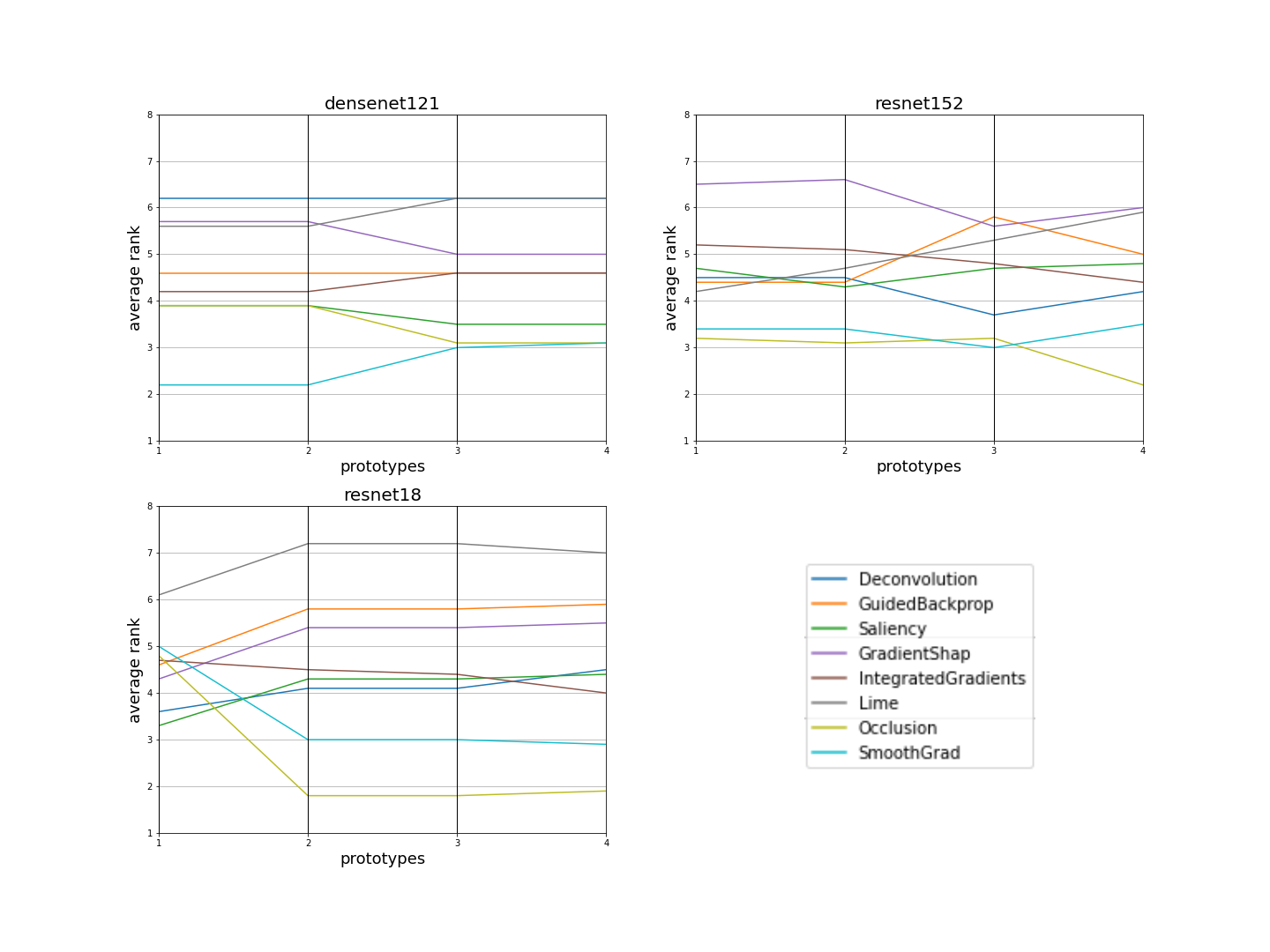}%
}
\caption[Average rank of saliency map methods based on prototypes.]{Plot showing the average rank of saliency map methods based on the four selected prototypes.}
\label{fig:PREPRINT_parallel}%
\end{figure}

Figure~\ref{fig:PREPRINT_parallel} shows the average rank of the saliency map method depending on the prototype.
Three interesting patterns emerge when we analyze the plot.
First, depending on the architecture chosen the overlap of the methods can vary greatly.
Second, despite this variation, some methods seem to produce saliency maps with a bigger overlap to ProtoPNet regardless of the architecture (i.e. SmoothGrad and Occlusion) while others show a smaller overlap (i.e. Deconvolution and Lime).
Lastly, SmoothGrad saliency maps are less fine-grained than other back-propagation methods, so it makes sense because the attribution maps created by ProtoPnet have a low resolution and are upscaled to the input image size.

\section{Conclusions}
\label{sec:Conclusions}

In this work, we proposed an approach to validate saliency map methods by measuring their overlap with the attribution maps produced by the intrinsic interpretable model ProtoPNet. As ProtoPNet does not use an external method to generate the attribution map it can be used as a ground-truth for the post-hoc methods.
In our experimental setup, we compared 8 different popular post-hoc saliency map methods and the prototypical attribution maps generated by ProtoPnet. This was performed by looking at the closeness of predicted labels and measuring the overlap of saliency maps with ProtoPNet, with 10 different saliency metrics adapted from literature on saliency models. 

ProtoPNet was not shown to trade-off performance in pursuit of interpretability, having achieved the most accurate model across all architectures. Also, the predictions of CNNs and ProtoPNets have been shown to correlate with each other.

While the saliency map methods produced a more fine-grained heatmap, ProtoPNet's attribution maps were soft and smooth. One possible justification relies on the prototype dimensions that the hyperparameter optimization chose, which were small. Also, we did not use fine annotation, as used in the IAIA-BL extension~\cite{Cynthia2021}, as they were not available in the Camelyon16.

Overall, in spite of some differences in results depending on the architecture chosen, two methods have been found to have statistically a bigger overlap with ProtoPNet: SmoothGrad and Occlusion. Deconvolution and Lime have shown consistently lower overlap. One possible reason for these results is the fact that ProtoPNet produces smooth attribution maps. While not as smooth as ProtoPNet, SmoothGrad, and Occlusion are more sparse and conservative in the highlighting, while Deconvolution and Lime are more fine-grained but also more dispersed.

In this work, we focused on the saliency maps produced by the last fully-connected layer of the network. Further extensions must compare salience maps from intermediate layers. By doing so we can ascertain if the feature being detected by a filter or neuron of the network is related to the prototypes.  

Future work directions also include the validation of the results on a dataset with fine annotation. An extension of ProtoPNet, called IAIA-BL\cite{Cynthia2021}, is capable of producing more fine-grained attribution maps by feeding the network fine annotation given by pathologists. While annotations are not always available, there are many public datasets where they are.

\bibliography{mybibfile}

\end{document}

%% file: tables/performance.tex
\begin{tabular}{llcccc}
\toprule
model &  base\_model &             auc &        accuracy &       precision &          recall \\
\midrule
PPNet & densenet121 & \textbf{0.9814} & \textbf{0.9814} &          0.9846 & \textbf{0.9780} \\

  CNN & densenet121 &          0.9750 &          0.9750 & \textbf{0.9869} &          0.9627 \\
PPNet &    resnet18 &          0.9703 &          0.9702 &          0.9667 &          0.9740 \\
  CNN &    resnet18 &          0.9780 &          0.9780 &          0.9792 &          0.9768 \\
PPNet &   resnet152 &          0.9545 &          0.9545 &          0.9444 &          0.9658 \\
  CNN &   resnet152 &          0.9628 &          0.9628 &          0.9647 &          0.9606 \\
\bottomrule
\end{tabular}

%% file: tables/labels.tex
\begin{tabular}{lccccc}
\toprule
base\_model &              auc &         accuracy &        precision &           recall \\
\midrule
densenet121 &  \textbf{0.9749} &  \textbf{0.9747} &  \textbf{0.9656} &  \textbf{0.9831} \\
resnet18 &           0.9703 &           0.9702 &           0.9654 &           0.9752 \\
resnet152 &           0.9540 &           0.9539 &           0.9418 &           0.9672 \\
\bottomrule
\end{tabular}

%% file: tables/average_rank.tex
\begin{tabular}{|l|cccc|cccc|cccc|}
\hline
 & \multicolumn{4}{|c|}{densenet121} & \multicolumn{4}{|c|}{resnet152} & \multicolumn{4}{|c|}{resnet18} \\
saliency\_method &    1 &    2 &    3 &    4 &  1 &    2 &    3 &    4 &  1 &    2 &    3 &    4\\
\hline
Deconvolution &  \textbf{6.2} &  \textbf{6.2} &  \textbf{6.2} &  \textbf{6.2} & 3.6 &  4.1 &  4.1 &  4.5 & 4.5 &  4.5 &  3.7 &  4.2\\
GuidedBackprop &  4.6 &  4.6 &  4.6 &  4.6 & 4.6 &  5.8 &  5.8 &  5.9 & 4.4 &  4.4 &  \textbf{5.8} &  5.0\\
Saliency &  3.9 &  3.9 &  3.5 &  3.5 & 3.3 &  4.3 &  4.3 &  4.4 &  4.7 &  4.3 &  4.7 &  4.8\\
GradientShap &  5.7 &  5.7 &  5.0 &  5.0 & 4.3 &  5.4 &  5.4 &  5.5 & \textbf{6.5} &  \textbf{6.6} &  5.6 &  6.0\\
IntegratedGradients &  4.2 &  4.2 &  4.6 &  4.6 & 4.7 &  4.5 &  4.4 &  4.0 & 5.2 &  5.1 &  4.8 &  4.4\\
Lime &  \textbf{5.6} &  \textbf{5.6} &  \textbf{6.2} &  \textbf{6.2} & \textbf{6.1} &  \textbf{7.2} &  \textbf{7.2} &  \textbf{7.0} & 4.2 &  4.7 &  5.3 &  \textbf{5.9}\\
Occlusion &  3.9 &  3.9 &  3.1 &  3.1 & 4.8 &  1.8 &  1.8 &  1.9 & 3.2 &  3.1 &  3.2 &  2.2\\
SmoothGrad &  2.2 &  2.2 & 3.0 & 3.1 & 5.0 &  3.0 &  3.0 &  2.9 & 3.4 &  3.4 &  3.0 &  3.5\\
\bottomrule
\end{tabular}